\documentclass[10pt,twocolumn,letterpaper]{article}

\usepackage[]{cvpr}
\usepackage{times}
\usepackage{epsfig}
\usepackage{graphicx}
\usepackage{amsmath}
\usepackage{amssymb}
\usepackage{booktabs}
\usepackage{multirow}
\usepackage{array}
\usepackage{microtype}
\usepackage{xcolor}
\usepackage{tikz}
\usetikzlibrary{shapes.geometric, arrows.meta, positioning, fit, backgrounds, calc}

\begin{document}

\title{GAFSV-Net: A Vision Framework for Online Signature Verification}

\author{
  Himanshu Singhal$^{1}$ \quad Suresh Sundaram$^{1}$ \\
  $^{1}$ Indian Institute of Technology Guwahati\\
  \small\texttt{\{h.singhal, sureshsundaram\}@iitg.ac.in}
}

\maketitle
\thispagestyle{empty}

\begin{abstract}
Online signature verification (OSV) requires distinguishing skilled forgeries from genuine samples under high intra-class variability and with very few enrollment samples. Existing deep learning methods operate directly on raw temporal sequences, restricting them to 1D architectures and preventing the use of pretrained 2D vision backbones. We bridge this gap with \textbf{GAFSV-Net}, which represents each signature as a six-channel asymmetric Gramian Angular Field image: three kinematic channels (pen speed, pressure derivative, direction angle) are each encoded into complementary GASF and GADF matrices that capture pairwise temporal co-occurrence and directional transition structure respectively. A dual-branch ConvNeXt-Tiny encoder processes GASF and GADF independently, with bidirectional cross-attention enabling each branch to query discriminative patterns from the other before metric-space projection. Training uses semi-hard triplet loss with skilled-forgery hard-negative injection; verification is performed via cosine similarity against a small enrollment prototype. We evaluate on DeepSignDB and BiosecurID, outperforming all sequence-based baselines trained under identical objectives, demonstrating that the representational gain of 2D temporal encoding is consistent and independent of training procedure, with ablations characterising each design choice's contribution.

\end{abstract}

\vspace{-5mm}
\section{Introduction}

In this paper, we present  a novel end-to-end deep learning  framework that establishes the veracity of a user via their online signatures. During test time, based on the learned  model,  a decision is made to either 
accept the claimed signature as genuine or reject it as a forgery. Online signature verification leverages the dynamic characteristics of the handwriting process, such as pen trajectory, velocity, pressure, and timing, to authenticate users. \\ \indent
Prior work followed three broad paradigms. Classical approaches relied on handcrafted temporal features with distance-based or statistical classifiers such as Dynamic Time Warping (DTW) and Gaussian Mixture Models (GMMs)~\cite{7820063}, which required domain-engineered features and stored templates that generalize poorly across writing styles. Deep learning shifted toward automatic feature extraction: CNN-RNN hybrids~\cite{DBLP:journals/corr/abs-2002-10119,9787558} and LSTM/GRU models~\cite{8259229,8270004} improved temporal modeling but their sequential processing limits global pairwise correlation capture. Siamese CNNs with contrastive or triplet objectives~\cite{sekhar2019osvnetconvolutionalsiamesenetwork} achieved competitive results, yet all these approaches process raw 1D temporal sequences directly, limiting their ability to leverage 2D spatial inductive biases and ImageNet-pretrained vision backbones.
\\ \indent
Transformer architectures~\cite{vaswani2017attention} have recently been applied
  to signature verification: OSVConTramer~\cite{10449120} combines CNN and Transformer
  streams with strong results.  Yet all these approaches operate directly on raw 1D temporal sequences,
precluding the use of ImageNet-pretrained 2D vision backbones and the rich spatial inductive biases they encode. 
\\ \indent
Structured 2D image representations offer a cleaner route: Gramian Angular Fields (GAF)~\cite{wang2015encoding} encode the full pairwise angular relationships between time-series values as a square matrix image in which every pixel captures the interaction between two time steps. Two complementary variants encode different aspects of this structure: GASF records cosine sums of angle pairs, capturing value co-occurrence, while GADF records sine differences, capturing directional transitions, as illustrated in Figure~\ref{fig:gaf}. Despite strong results in general time-series classification~\cite{wang2015encoding,hatami2018classification}, neither variant has been explored for OSV. We address this gap with \textbf{GAFSV-Net}, a dual-branch encoder that represents each signature as a six-channel GASF+GADF image of three kinematic channels and trains a writer-independent metric-learning framework on the resulting representations.
 
\medskip\noindent\textbf{Contributions:}
\vspace{-2mm}
\begin{enumerate}
  \item We propose GAF-based encoding of signature time series that (i) renders pairwise temporal correlations spatially explicit and (ii) enables transfer of ImageNet-pretrained vision backbones to a data-scarce domain.

  \vspace{-2mm}
  \item We show that GASF and GADF capture orthogonal aspects of temporal dynamics (value co-occurrence vs. directional transitions), with each channel exposing forgery artifacts the other cannot.
  \vspace{-2mm}
  \item We present the first controlled, representation-focused comparison between GAF-based and sequence-based methods for OSV, holding the training objective fixed across a diverse set of sequence encoders.
\end{enumerate}

\section{Related Work}

Classical approaches to OSV relied on handcrafted temporal features with Dynamic Time Warping (DTW)~\cite{7820063} or Hidden Markov Model (HMM) classifiers~\cite{KHOLMATOV20052400,10.5555/1762222.1762254,FIERREZ20072325}, operating on features such as position, velocity, pen pressure, and stroke statistics~\cite{4603099}. While effective on constrained datasets, these methods required domain-engineered feature design and stored reference templates, limiting generalization across writing styles and acquisition conditions~\cite{FIERREZ20072325}.
\\ \indent
Deep learning enabled automatic feature extraction from raw signatures. RNN-based models (LSTMs, GRUs)~\cite{8259229,8270004} demonstrated strong temporal modeling but their sequential processing limits global pairwise correlation capture, as hidden states primarily encode local context. DeepSign~\cite{DBLP:journals/corr/abs-2002-10119} achieved strong results with CNN-RNN hybrids, inverse discriminative networks~\cite{8954001} jointly learned features and decision boundaries, and the DsDTW framework~\cite{9787558} integrated differentiable soft-DTW into a convolutional recurrent architecture, winning the ICDAR 2021 competition. Siamese CNNs with contrastive or triplet objectives~\cite{sekhar2019osvnetconvolutionalsiamesenetwork} have shown competitive performance, though gradient backpropagation through separate branches can lead to asymmetric feature learning across genuine and impostor manifolds.
\\ \indent 
Transformer architectures~\cite{vaswani2017attention} have been applied to signature verification, but continuous temporal sequences pose challenges for positional encodings designed for discrete tokens~\cite{melzi2023exploringtransformersonlinehandwritten}. OSVConTramer~\cite{10449120} addresses this through hybrid CNN-Transformer streams. Nevertheless, all these approaches operate on raw 1D sequences or rendered trajectory images, precluding the use of ImageNet-pretrained 2D vision backbones.
\\ \indent 
Gramian Angular Fields (GAF)~\cite{wang2015encoding} transform temporal signals into 2D matrix images by encoding pairwise angular relationships in polar space. The two complementary variants  GASF (cosine of summed angles) and GADF (sine of difference angles) have been successfully applied to general time-series classification~\cite{hatami2018classification}, enabling 2D convolutional architectures to exploit spatial inductive biases. Neither variant has been applied to OSV.
\\ \indent
Writer-independent verification remains challenging, requiring discriminative features that generalize across writers and acquisition conditions~\cite{FIERREZ20072325}, particularly under limited enrollment. We bridge this gap with a dual-branch encoder that processes GASF and GADF views of three kinematic channels jointly through bidirectional cross-attention, trained via metric learning with semi-hard triplet loss, skilled-forgery hard-negative injection, and a uniformity regulariser, without user-specific parameters at test time.




\vspace{-2mm}
\section{Overview of Our Method}
\label{sec:overview}
\vspace{-1mm}
Our framework processes a raw $(x,y,p)$ stylus sequence through four successive stages. First, three kinematic time series are extracted: pen speed $v$, pressure derivative $\dot{p}$, and direction angle $\theta$ (Section~\ref{sec:gaf}). Each series is independently resampled and encoded into a complementary pair of 2D images, a Gramian Angular Summation Field (GASF) and a Gramian Angular Difference Field (GADF), yielding a single six-channel image. Second, the six-channel image is split into GASF and GADF halves, each processed by a dedicated ConvNeXt-Tiny backbone; intra-branch spatial self-attention refines the resulting feature maps, and bidirectional cross-attention fuses information across the two branches (Section~\ref{sec:arch}). Third, the fused representation is projected onto the unit hypersphere, producing a compact $D$-dimensional embedding. Finally, the model is trained writer-independently with a semi-hard triplet loss augmented by skilled-forgery hard-negative injection and a uniformity regulariser (Section~\ref{sec:loss}). At test time, verification reduces to a cosine similarity comparison between a query embedding and the mean of a writer's reference embeddings (Section~\ref{sec:inference}).

\begin{figure}[!t]
  \centering
  \includegraphics[width= 0.8 \columnwidth]{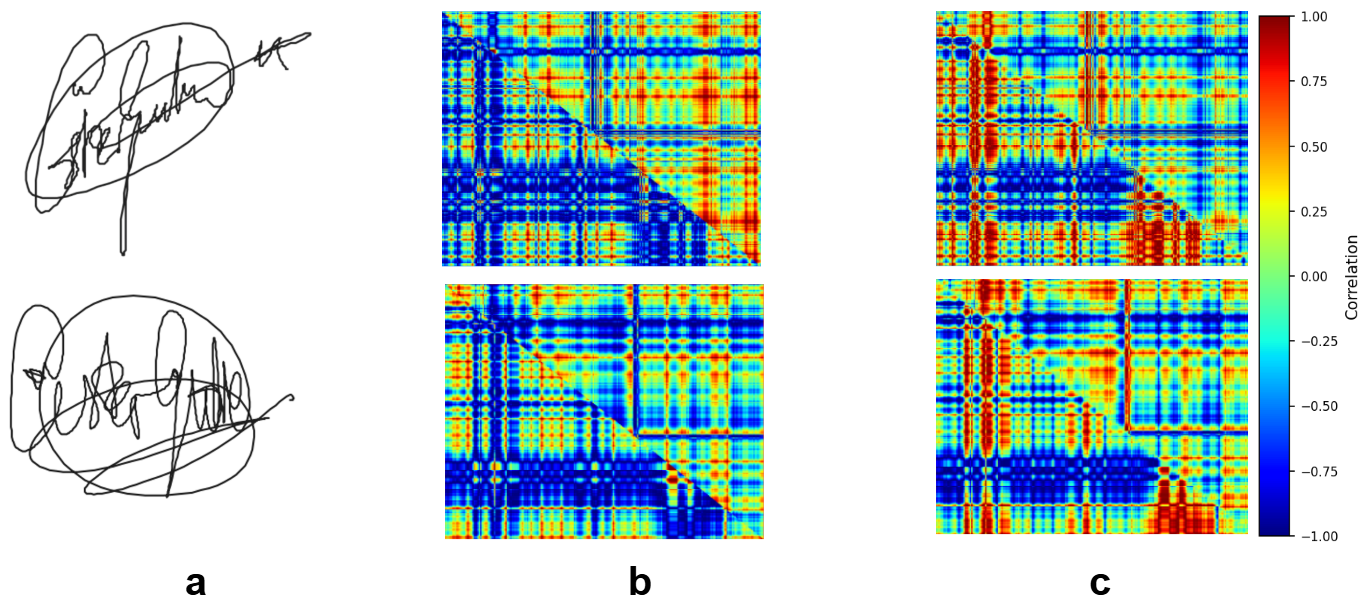}
  \caption{Raw signature trajectories (a), Asymmetric GASF (b), and GADF (c) images from the speed channel for a genuine (top row) and a skilled forgery (bottom row) from the same writer in DeepSignDB. GASF's symmetric structure captures value co-occurrence; GADF's anti-symmetric structure with zero diagonal captures directional transitions. Both representations exhibit systematic differences between genuine and forged samples, supporting their complementarity as dual-branch inputs.}
  \label{fig:gaf}
\end{figure}

\begin{figure*}[!ht]
  \centering
  \includegraphics[width=\textwidth]{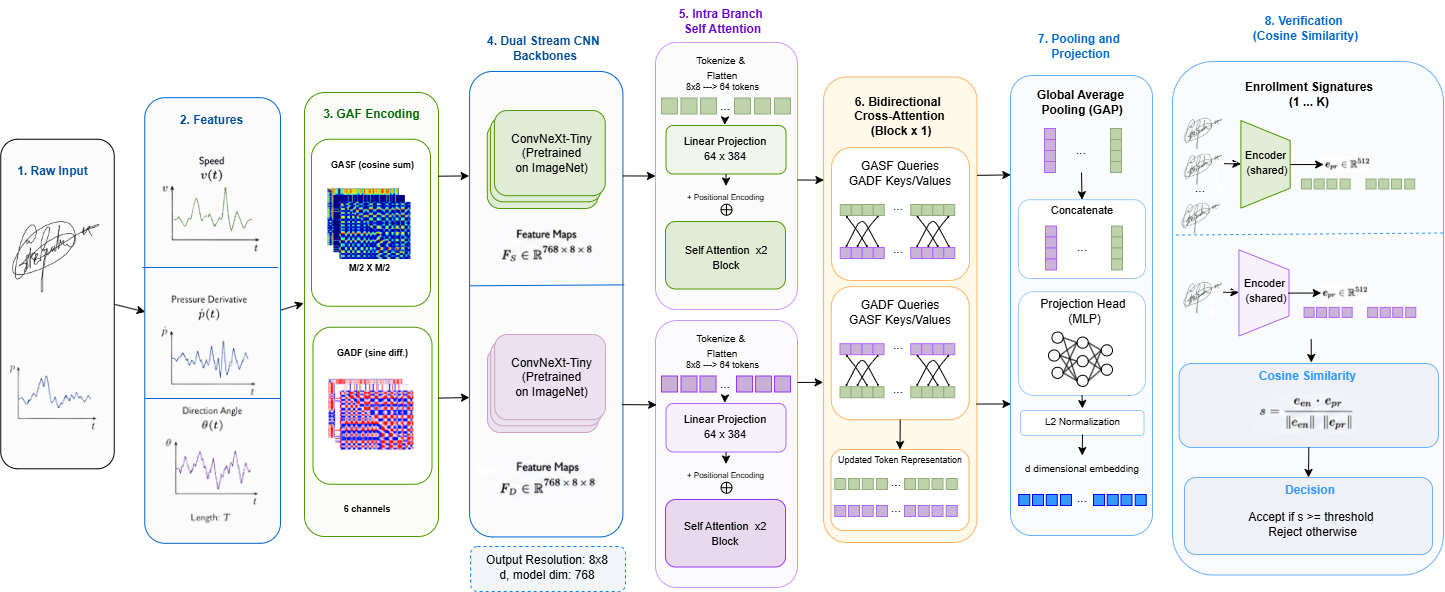}
  \caption{\textbf{Overview of the proposed dual-branch architecture for online signature verification.} Three kinematic signals: speed $v(t)$, pressure derivative $\dot{p}(t)$, and direction angle $\theta(t)$, are encoded into GASF and GADF images, forming two 3-channel inputs processed by independent ConvNeXt-Tiny backbones. The resulting feature maps are tokenized, refined through intra-branch self-attention, and fused via bidirectional cross-attention. Global average pooling of both branches is concatenated and projected to a $d$-dimensional embedding on the unit hypersphere. At verification, a query embedding is scored against the mean enrollment prototype by cosine similarity and accepted if the score exceeds a threshold $\tau$.}
  \label{fig:arch}
\end{figure*}

\section{Gramian Angular Fields}
\label{sec:gaf}
Online signatures from stylus devices provide $(x,y,p)$ sequences of position and pen pressure. We extract three time series via central-difference derivatives:
\begin{align}
v(t) &= \sqrt{v_x(t)^2 + v_y(t)^2}, \\
\dot{p}(t) &= \frac{dp}{dt}, \\
\theta(t) &= \tan^{-1} \frac{v_y} {v_x}.
\end{align}
Each time series is independently resampled to $M$ points and normalised to $[-1,1]$, then encoded into asymmetric GASF and GADF images ($\frac{M}{2}{\times}\frac{M}{2}$) and stacked into:

\begin{equation}
\mathbf{I} = \scalebox{0.9}{$\displaystyle\bigl[
G^{(v)\text{S}}_\text{asym},\; G^{(v)\text{D}}_\text{asym},\;
G^{(\dot{p})\text{S}}_\text{asym},\; G^{(\dot{p})\text{D}}_\text{asym},\;
G^{(\theta)\text{S}}_\text{asym},\; G^{(\theta)\text{D}}_\text{asym}
\bigr] \in \mathbb{R}^{6\times \frac{M}{2}\times \frac{M}{2}}$}.
\label{eq:6ch}
\end{equation}

\textbf{Gramian Angular Summation (GASF)}: Let $\tilde{X} = (\tilde{x}_1, \ldots, \tilde{x}_M)$ be a normalised univariate time series. Each value is phase-encoded:
\begin{equation}
  \phi_i = \arccos(\tilde{x}_i), \quad \tilde{x}_i \in [-1, 1].
  \label{eq:phi}
\end{equation}
The GASF matrix is:
\begin{equation}
  G^\text{S}[i,j] = \cos(\phi_i + \phi_j)
           = \tilde{x}_i\tilde{x}_j - \sqrt{1-\tilde{x}_i^2}\sqrt{1-\tilde{x}_j^2}.
  \label{eq:gasf}
\end{equation}
$G^\text{S}$ is \emph{symmetric}: $G^\text{S}[i,j]=G^\text{S}[j,i]$. Its diagonal encodes each individual value; off-diagonal entries encode pairwise temporal correlations. Large values occur when both $\phi_i$ and $\phi_j$ are small (both values near $+1$) or both near $\pi$ (both near $-1$), making $G^\text{S}$ an image of \emph{value co-occurrence} in phase space.

 \textbf{Gramian Angular Difference (GADF)} : The GADF matrix is:
\begin{equation}
  G^\text{D}[i,j] = \sin(\phi_i - \phi_j)
    = \sqrt{1-\tilde{x}_i^2}\,\tilde{x}_j - \tilde{x}_i\sqrt{1-\tilde{x}_j^2}.
  \label{eq:gadf}
\end{equation}
$G^\text{D}$ is \emph{anti-symmetric}: $G^\text{D}[i,j] = -G^\text{D}[j,i]$, with an \emph{identically zero diagonal}. $G^\text{D}[i,j] > 0$ when $\tilde{x}_j > \tilde{x}_i$, so GADF encodes the \emph{direction of temporal change} between time steps. The two matrices are therefore structurally complementary: GASF captures value co-occurrence in phase space while GADF captures directional transitions, and features prominent in one are suppressed or absent in the other by construction.

\textbf{Asymmetric Construction}: A full $M{\times}M$ GAF matrix yields images that are unnecessarily large and redundant. We adopt an \emph{asymmetric} construction: the series is split at the midpoint, and each half is independently converted to a $\frac{M}{2}{\times}\frac{M}{2}$ matrix. Crucially, a single normalisation pass is applied to the \emph{complete} series before the split, so that the relative magnitude between the early and late phases of the signature is preserved across both halves. The asymmetric image combines the two halves of the matrices $  G_1[i,j]$ and $  G_2[i,j]$
\begin{equation}
  G_\mathrm{asym}[i,j] = \begin{cases}
    G_1[i,j] & i \leq j \\
    G_2[i,j] & i > j
  \end{cases}
  \label{eq:asym}
\end{equation}
For a sequence of length $M$ this yields a $\frac{M}{2}{\times}\frac{M}{2}$ image. The upper triangle encodes early-phase correlations; the lower triangle encodes completion-phase correlations. This construction applies identically to both GASF and GADF, preserving their respective structural properties (symmetry for GASF, anti-symmetry for GADF) within each triangle.

Figure~\ref{fig:gaf} illustrates the asymmetric GASF and GADF images for a genuine and forged signature, highlighting their complementary visual structures.

For OSV, this complementarity is practically meaningful: GADF applied to pen speed exposes the precise sequence of accelerations and decelerations that a forger replicating the gross speed profile will fail to reproduce, while GADF on the pressure derivative directly reveals whether pressure is rising or falling between any two time steps — inconsistencies that GASF may partially obscure through its averaging effect. Because GASF and GADF have fundamentally different spatial statistics, a single shared backbone is ill-suited to both, motivating separate branches with independently learned weights.

\section{Dual-Branch Framework}
\label{sec:method}

\subsection{Architecture}
\label{sec:arch}

Figure~\ref{fig:arch} gives an overview of the full pipeline. The six-channel input $\mathbb{R}^{6\times \frac{M}{2}\times \frac{M}{2}}$ is split into GASF and GADF sub-inputs, each comprising three channels:
\[
  \mathbf{I}^\text{S} = \mathbf{I}_{[0,2,4]} \in \mathbb{R}^{3\times\frac{M}{2}\times\frac{M}{2}}, \quad
  \mathbf{I}^\text{D} = \mathbf{I}_{[1,3,5]} \in \mathbb{R}^{3\times\frac{M}{2}\times\frac{M}{2}}.
\]

\textbf{Dual ConvNeXt backbones.}  Two independent ConvNeXt-Tiny~\cite{liu2022convnet} networks pretrained on ImageNet-1k process each sub-input with classifier heads and global pooling removed, producing feature maps of spatial size $S{\times}S$ and channel dimension $D_\text{bb}$:
\begin{align}
  \mathbf{F}^\text{S} &= \mathrm{BB}_\text{GASF}(\mathbf{I}^\text{S}) \;\in\; \mathbb{R}^{D_\text{bb}\times S\times S}, \\
  \mathbf{F}^\text{D} &= \mathrm{BB}_\text{GADF}(\mathbf{I}^\text{D}) \;\in\; \mathbb{R}^{D_\text{bb}\times S\times S}.
\end{align}
Separate backbones allow each to develop filters specialised for its transform's visual statistics: symmetric images with structured diagonals (GASF) versus anti-symmetric images with zero diagonals (GADF).

\textbf{Token projection and positional embedding.} Spatial feature maps are flattened into $n_\text{tok}$ token sequences and projected to a shared $d$-dimensional space via branch-specific linear layers. A shared learned positional embedding $\mathbf{e}_\text{pos}\!\in\!\mathbb{R}^{n_{tok}\times d}$ is added to both:
\begin{eqnarray}
  \mathbf{H}^\text{S} = \mathbf{W}^\text{S}\,\mathrm{flat}(\mathbf{F}^\text{S}) + \mathbf{e}_\text{pos}, \quad \\ 
  \mathbf{H}^\text{D} = \mathbf{W}^\text{D}\,\mathrm{flat}(\mathbf{F}^\text{D}) + \mathbf{e}_\text{pos}.
\end{eqnarray}

\textbf{Intra-branch self-attention.} Two multi-head self-attention layers~\cite{vaswani2017attention} with feed-forward sublayers are applied independently in each branch, capturing global spatial correlations within each GAF representation:
\begin{eqnarray}
  \mathbf{H}^{\text{S}\prime} = \mathrm{SelfAttn}^{(2)}_\text{GASF}(\mathbf{H}^\text{S}), \quad \\
  \mathbf{H}^{\text{D}\prime} = \mathrm{SelfAttn}^{(2)}_\text{GADF}(\mathbf{H}^\text{D}).
\end{eqnarray}

\textbf{Bidirectional cross-attention.} A single bidirectional cross-attention block enables inter-branch fusion. Both directions are computed in parallel from the pre-update tokens, so neither branch's update affects the other's keys/values within the same block:
\begin{align}
  \tilde{\mathbf{H}}^\text{S} &= \mathbf{H}^{\text{S}\prime} + \mathrm{MHA}(\mathbf{H}^{\text{S}\prime},\; \mathbf{H}^{\text{D}\prime},\; \mathbf{H}^{\text{D}\prime}), \label{eq:cross_a}\\
  \tilde{\mathbf{H}}^\text{D} &= \mathbf{H}^{\text{D}\prime} + \mathrm{MHA}(\mathbf{H}^{\text{D}\prime},\; \mathbf{H}^{\text{S}\prime},\; \mathbf{H}^{\text{S}\prime}), \label{eq:cross_b}
\end{align}
followed by residual FFNs in each branch. Eq.~\ref{eq:cross_a} allows the GASF branch to  query which spatial tokens of the GADF representation are most discriminative. Likewise,  Eq.~\ref{eq:cross_b}  swaps the role of   GADF branch to  query  and the key-value to GASF.

\textbf{Aggregation and projection head.} Post-norm global average pooling produces per-branch vectors that are concatenated:
\begin{equation}
  \mathbf{v} = \bigl[\mathrm{GAP}(\mathrm{LN}(\tilde{\mathbf{H}}^\text{S}));\;\mathrm{GAP}(\mathrm{LN}(\tilde{\mathbf{H}}^\text{D}))\bigr] \in \mathbb{R}^{2d}.
\end{equation}
A BN-projection head~\cite{chen2020simple} maps $\mathbf{v}$ to a $d_z$-dimensional unit-norm embedding:
\begin{equation}
  \mathbf{z} = \mathrm{L2}\bigl(f_2\bigl(\mathrm{ReLU}(\mathrm{BN}(f_1(\mathbf{v})))\bigr)\bigr),
  \label{eq:proj}
\end{equation}
where $f_1\!:\!\mathbb{R}^{2d}\!\to\!\mathbb{R}^{2d}$ (no bias) and $f_2\!:\!\mathbb{R}^{2d}\!\to\!\mathbb{R}^{d_z}$.

\subsection{Training Objective}
\label{sec:loss}

Each training step samples $W_b$ writers, each contributing $R{+}1$ genuine
images, yielding $N_g$ genuine L2-normalized embeddings ${\mathbf{z}_i} \subset \mathbb{R}^D$, with writer labels ${y_i}$.
An additional $B_f$ skilled forgery images are encoded to embeddings
$\{\mathbf{z}^f_k\}$, each assigned a unique label not shared with any genuine sample.
This makes forgeries available as hard-negative candidates during mining without
ever being anchors or positives themselves.
 \begin{figure}[!ht]
  \centering
  \includegraphics[width= 1 \columnwidth]{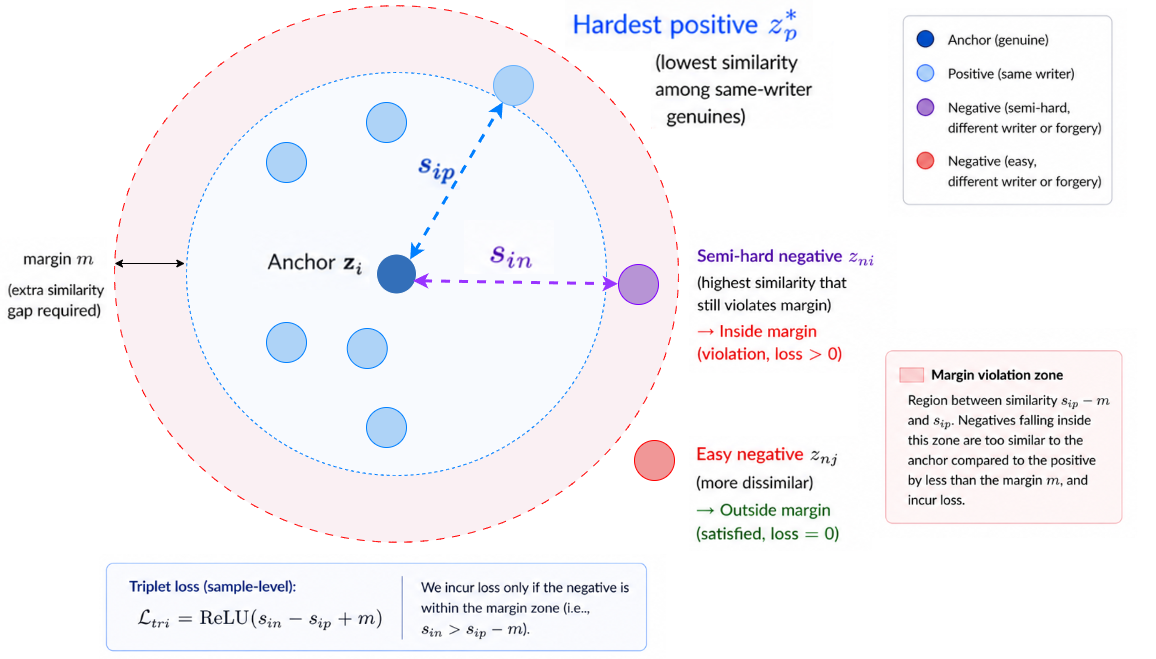}
  \caption{Illustration of the semi-hard triplet loss, where loss is incurred only when a negative falls within the margin violation zone (i.e., $s_{in} > s_{ip} - m$).}
  \label{fig:loss}
\end{figure}

\vspace{-5mm}
\paragraph{Triplet loss with forgery injection.}
  For each genuine anchor $\mathbf{z}_i$, we mine the hardest positive $\mathbf{z}_{p^*_i}$
  (lowest cosine similarity among same-writer genuines) and a semi-hard negative
  $\mathbf{z}_{n_i}$ from the combined pool of genuines and forgeries: the
  highest-cosine different-writer sample that still lies within the margin violation
  zone ($\mathbf{z}_i\!\cdot\!\mathbf{z}_{p^*_i} - m < \mathbf{z}_i\!\cdot\!\mathbf{z}_{n_i} < \mathbf{z}_i\!\cdot\!\mathbf{z}_{p^*_i}$),
  falling back to the hardest negative when none exists.
  This operates at the \emph{sample level}.
  However, sample-level mining cannot enforce that a forgery is specifically rejected
  by its \emph{target} writer's cluster.
  We therefore add a \emph{cluster-level} term: for each forgery $\mathbf{z}^f_k$ of
  writer $w^f_k$, let $\bar{s}^\text{pos}$ be the mean intra-cluster genuine similarity
  and $\bar{s}^\text{neg}_k$ the mean similarity of that writer's genuines to the forgery. Figure~\ref{fig:loss} illustrates the margin violation zone and the roles of the hardest positive and semi-hard negative in this formulation.
  The combined objective enforces a margin $m$ between positive and negative similarities, with ReLU ensuring zero loss on already-satisfied triplets:

\begin{equation}
\begin{aligned}
\mathcal{L}_\text{tri} =
&\;\underbrace{\frac{1}{|\mathcal{A}|} \sum_{i \in \mathcal{A}}
  \mathrm{ReLU}\bigl(
    \mathbf{z}_i \cdot \mathbf{z}_{n_i} - \mathbf{z}_i \cdot \mathbf{z}_{p^*_i} + m
  \bigr)}_{\text{sample-level: global writer separation}} \\
&\;+\;
\frac{\lambda_f}{|\mathcal{K}|} \underbrace{\sum_{k \in \mathcal{K}}
  \mathrm{ReLU}\bigl(
    \bar{s}^\text{neg}_k - \bar{s}^\text{pos}_{w^f_k} + m
  \bigr)}_{\text{cluster-level: per-writer forgery rejection}}
\end{aligned}
\label{eq:triplet}
\end{equation}
  where $\mathcal{A}$ is the set of genuine anchors with at least one positive,
  $\mathcal{K}$ contains forgeries whose writer has at least two genuine embeddings
  in the batch.

  \paragraph{Uniformity regulariser.}
  Triplet losses optimise only relative distances and can collapse all embeddings
  into a small region of the sphere. The uniformity loss~\cite{wang2020understanding}
  counteracts this by encouraging all $N{=}N_{g}+B_{f}$ embeddings to spread evenly over
  the embedding space:
  \begin{equation}
    \mathcal{L}_\text{unif} = \log \frac{1}{N^2}
      \sum_{i,j} e^{-\frac{\|\mathbf{z}_i - \mathbf{z}_j\|^2}{2}}.
    \label{eq:unif}
  \end{equation}

  The total loss is:
  \begin{equation}
    \mathcal{L} = \mathcal{L}_\text{tri}
                + \lambda_u\,\mathcal{L}_\text{unif}
    \label{eq:total}
  \end{equation}
  $\mathcal{L}_\text{tri}$ combines two complementary objectives: its sample-level
  term separates all writers globally, while its cluster-level forgery term directly
  tightens the decision boundary between each writer's genuine cluster and their
  known forgeries. $\mathcal{L}_\text{unif}$ ensures the full capacity of the
  embedding space is utilised. Ablation studies (Section~\ref{sec:analysis})
  confirm the complementary role of each component.


\subsection{Verification Scoring}
\label{sec:inference}
At test time, $R_\text{enroll}$ reference embeddings $\{\mathbf{z}_{r_k}\}$ are averaged to produce an aggregate reference embedding $\bar{\mathbf{z}}_r = \frac{1}{R_\text{enroll}}\sum_k \mathbf{z}_{r_k}$. The verification score for a query sample $q$ is then computed as the cosine similarity $s(q) = \mathbf{z}_q \cdot \bar{\mathbf{z}}_r$, and the signature is accepted if $s(q)$ exceeds a threshold $\tau$, else it is rejected as forgery. 

\section{Experimental Protocol with implementation details}
\label{sec:experiments}
We evaluate on two publicly available benchmarks.

\textbf{DeepSignDB}~\cite{tolosana2021deepsigndb} is a large-scale composite stylus database comprising 1,526 contributors from multiple devices and sessions. We consider the stylus setting and follow the \emph{standard DeepSignDB evaluation protocol}: the Development set (574 users) for training, the Evaluation set (442 users) for testing. The Evaluation set includes both skilled ($\mathrm{sf}$) and random forgery ($\mathrm{rf}$) subsets evaluated separately.

\textbf{BiosecurID}~\cite{fierrez2010biosecurid} consists of 400 users across four sessions on a digitising tablet, with 16 genuine and 12 skilled forgeries per user. We adopt the standard session-independent evaluation protocol with $R_\text{enroll}{=}4$ reference samples.

We report \emph{Equal Error Rate} (EER), the operating threshold at which FRR equals FAR, computed globally over all evaluation pairs following the DeepSignDB protocol~\cite{tolosana2021deepsigndb}. Skilled and random forgeries are evaluated separately where applicable. We report results for enrollment sizes $R_\text{enroll}{\in}\{1,4\}$. \\ \indent
Six-channel GASF+GADF images are computed offline from raw signatures resampled to $M{=}512$ points and stored as \texttt{float16} arrays of size $256{\times}256$. Both ConvNeXt-Tiny backbones are initialised from ImageNet-1k weights via \texttt{timm}~\cite{rw2019timm}; all parameters are fine-tuned jointly. Batch size is 320 genuine + 64 forgery images per step. All experiments run on a single NVIDIA A100 (40\,GB)  GPU.

\section{Results}
\label{sec:results}

\subsection{Per-Dataset Performance}
\label{sec:per_dataset}

Table~\ref{tab:per_dataset} reports GAFSV-Net EER broken down across the five sub-datasets of DeepSignDB and BiosecurID, under both enrollment sizes.

Performance varies substantially across sub-datasets, reflecting differences in acquisition device, session variability, and forgery difficulty.
\textbf{BiosecurID} achieves the lowest skilled EER (2.97\% at $R_\text{enroll}{=}4$) despite containing 12 adversarial skilled forgeries per writer: the largest forgery set in any sub-dataset. The combination of writers across 4 acquisition sessions in training, a high-fidelity digitising tablet, and consistent pressure sampling produces clean $\dot{p}$ trajectories that sharply separate genuine from forged embeddings.
\textbf{MCYT} (5.57\%) benefits similarly from a high-quality Wacom tablet and low intra-writer variability; skilled forgeries in MCYT tend to reproduce gross shape but fail to match the fine velocity profile, which the GASF branch captures effectively.
\textbf{BiosecureDS2} (5.94\%) and \textbf{eBioSignDS2} (3.45\%) occupy the mid-range; both share the same eBioSign acquisition device and recording setup, yet eBioSignDS2 is notably easier, likely because its signatures were collected in a single controlled working condition, keeping intra-writer variability low.
\begin{table}[!ht]
  \centering
  \caption{GAFSV-Net EER (\%) per sub-dataset. sf\,=\,skilled, rf\,=\,random. $^\dagger$Global aggregate.}
  \label{tab:per_dataset}
  \small
  \setlength{\tabcolsep}{3.5pt}
  \renewcommand{\arraystretch}{1.05}
  \begin{tabular}{@{}lcccc@{}}
    \toprule
    & \multicolumn{2}{c}{$R_\text{enroll}{=}1$} & \multicolumn{2}{c}{$R_\text{enroll}{=}4$}\\
    \cmidrule(lr){2-3}\cmidrule(lr){4-5}
    Dataset & sf & rf & sf & rf\\
    \midrule
    MCYT              & 8.39 & 1.45 &  5.57 & 0.58\\
    BiosecurID        & 5.27 & 1.64 &  2.97 & 1.07\\
    BiosecureDS2      &10.71 & 3.10 &  5.94 & 1.72\\
    eBioSignDS2       &10.51 & 3.21 &  3.45 & 0.69\\
    eBioSignDS1       &15.46 & 3.78 & 11.55 & 1.87\\
    \midrule
    DeepSignDB$^\dagger$ & \textbf{9.41} & \textbf{2.35} & \textbf{6.01} & \textbf{1.25}\\
    \bottomrule
  \end{tabular}
\end{table}

\textbf{eBioSignDS1} is the most challenging sub-dataset (11.55\%), as its signatures span five different working conditions (different body postures and surfaces), introducing substantial intra-writer variability in pen speed and pressure that widens the genuine-score distribution and raises confusion with skilled forgeries.
The gap between $R_\text{enroll}{=}1$ and $R_\text{enroll}{=}4$ is widest on the harder sub-datasets (eBioSignDS1: 15.46\%$\rightarrow$11.55\%, BiosecureDS2: 10.71\%$\rightarrow$5.94\%), confirming that a richer enrollment prototype suppresses intra-writer noise more effectively where variability is high.
Random forgery EER remains consistently low across all sub-datasets ($\leq$3.78\%), indicating that the embedding space cleanly separates writers who make no attempt to replicate the target style; the remaining challenge lies entirely in skilled forgeries where imitators reproduce coarse trajectory structure but fail to match the fine-grained $\dot{p}$ and $\theta$ dynamics encoded in the GAF images.

\subsection{Comparison with Baselines}
\label{sec:comparison}

Table~\ref{tab:main_results} summarises GAFSV-Net against published methods and sequence-based baselines on DeepSignDB and BiosecurID.
Since prior OSV work has operated predominantly on raw temporal sequences, no image-based comparison exists in the literature, and recent methods \cite{vorugunti2023osvcontramer} are not directly comparable under the standard DeepSignDB protocol; we therefore establish our own sequence baselines by training five encoders (BiGRU, BiLSTM, Vanilla Transformer, 1D-CNN, TCN) on the same kinematic channels ($v$, $\dot{p}$, $\theta$) with the identical triplet and forgery injection objective, providing a controlled comparison of representations. Among published methods, GAFSV-Net (6.01\% at $R_\text{enroll}{=}4$) is competitive with DTW (6.14\%) overall but substantially outperforms it on BiosecurID (2.97\% vs.\ 10.43\%), where richer pressure dynamics provide a stronger discriminative signal than DTW's alignment-based distance.
DeepSign~\cite{DBLP:journals/corr/abs-2002-10119} (2.43\%) surpasses GAFSV-Net on DeepSignDB: it employs 23 hand-engineered time functions including second-order derivatives, stroke geometry, and trigonometric encodings, pre-aligned via DTW before RNN encoding, giving it a richer input signal and explicit temporal alignment compared to GAFSV-Net's three-channel encoding. GAFSV-Net recovers this gap on BiosecurID (2.97\% vs.\ 5.61\%), where the dominant discriminator is the overall pressure trajectory shape and the ImageNet-pretrained backbone extracts richer structural features from the clean $\dot{p}$ channel than a model trained from scratch on a small dataset.

\begin{table}[!t]
  \centering
  \caption{EER (\%) on DeepSignDB Evaluation set and BiosecurID under standard protocols. sf\,=\,skilled, rf\,=\,random; `--'\,=\,not reported. $^\dagger$\,trained with same objective as GAFSV-Net. MOMENT uses a backbone pretrained on a large time-series corpus~\cite{goswami2024moment}.}
  \label{tab:main_results}
  \small
  \setlength{\tabcolsep}{5pt}
  \renewcommand{\arraystretch}{1.1}
  \begin{tabular}{@{}lcccc c@{}}
    \toprule
    & \multicolumn{4}{c}{\textbf{DeepSignDB}} & \textbf{BiosecurID}\\
    \cmidrule(lr){2-5}\cmidrule(lr){6-6}
    Method & \multicolumn{2}{c}{$R_\text{enroll}{=}1$} & \multicolumn{2}{c}{$R_\text{enroll}{=}4$} & $R_\text{enroll}{=}4$\\
    \cmidrule(lr){2-3}\cmidrule(lr){4-5}
    & sf & rf & sf & rf & sf\\
    \midrule
    DTW~\cite{munich1999continuous}                   & 9.87 & 3.82 & 6.14 & 2.01 & 10.43\\
    DeepSign~\cite{DBLP:journals/corr/abs-2002-10119} & 4.12 & 1.65 & 2.43 & 0.91 & 5.61\\
    BiGRU$^\dagger$                                   & 17.89 & 5.77 & 14.07 & 3.76 & 10.48\\
    BiLSTM$^\dagger$                                  & 16.36 & 5.48 & 12.80 & 3.74 & 8.52\\
    Vanilla Transformer$^\dagger$                     & 13.20 & 4.89 &  9.42 & 2.90 & 5.93\\
    1D-CNN$^\dagger$                                  & 11.80 & 3.12 &  8.61 & 1.63 & 5.11\\
    TCN$^\dagger$~\cite{bai2018empirical}                                     & 11.51 & 3.39 &  8.50 & 1.96 & 4.48\\
    MOMENT$^\dagger$~\cite{goswami2024moment}                                 & 11.02 & 4.78 &  7.83 & 2.96 & 4.17\\
    \midrule
    \textbf{GAFSV-Net (ours)}                           & \textbf{9.41} & \textbf{2.35} & \textbf{6.01} & \textbf{1.25} & \textbf{2.97}\\
    \bottomrule
  \end{tabular}
\end{table}

Against the controlled sequence baselines, GAFSV-Net outperforms every 1D encoder consistently across both datasets.
Recurrent models (BiGRU, BiLSTM) perform worst: compressing a 512-step sequence into a fixed vector under triplet gradients provides insufficient signal for subtle pressure and directional dynamics.
The Vanilla Transformer ranks below both convolutional baselines despite global self-attention: even with the full pairwise attention capacity of a transformer, processing kinematic signals as 1D token sequences cannot replicate the explicit 2D correlation structure that GAF encoding makes directly accessible to pretrained convolutional features. TCN~\cite{bai2018empirical} (8.50\% / 4.48\%) and 1D-CNN (8.61\% / 5.11\%) are the strongest sequence models yet still trail GAFSV-Net (6.01\% / 2.97\%) by 2.49\% and 1.51\% on DeepSignDB and BiosecurID respectively, as neither can exploit the cross-temporal co-occurrence structure that the 2D GAF encoding makes directly accessible to pretrained convolutional features.

MOMENT~\cite{goswami2024moment} (7.83\% / 4.17\%), a time-series foundation model pretrained on diverse sequence data (ECG, EEG, weather, clinical), still underperforms GAFSV-Net by 1.82\% on DeepSignDB and 1.20\% on BiosecurID under identical training. This isolates the contribution of the GAF representation: 2D encoding of pairwise temporal correlations provides inductive biases that sequence-domain pretraining does not recover, confirming GAFSV-Net's gains are not attributable solely to ImageNet transfer.

The advantage is consistent across both enrollment sizes, with the gap widening at $R_\text{enroll}{=}1$, indicating that GAF embeddings form a more stable prototype from fewer references.

\section{Analysis and Ablation}
\label{sec:analysis}

\subsection{Input Representation and Branch Configuration}
\label{sec:abl_branch}

Table~\ref{tab:ablation_branch} ablates the input representation and branch structure on DeepSignDB ($R_\text{enroll}{=}4$, 150 training epochs throughout). \emph{GASF-only} and \emph{GADF-only} are single-backbone models processing only the three GASF or GADF channels respectively, with the same ConvNeXt-Tiny backbone and self-attention head used in each branch of the full model. \emph{Trajectory image} replaces the GAF encoding with a single channel image (binary pen-down mask at 256$\times$256), providing a direct image-baseline that preserves 2D spatial shape without any temporal encoding. \emph{Dual, concat} uses both branches but replaces bidirectional cross-attention with direct vector concatenation. \emph{Dual, cross-attn} is the full proposed model.

\begin{table}[t]
  \centering
  \caption{Input representation and branch ablation, DeepSignDB. `$^*$' best, `$^{**}$' second best.}
  \label{tab:ablation_branch}
  \small
  \setlength{\tabcolsep}{4pt}
  \begin{tabular}{@{}lcc@{}}
    \toprule
    Configuration                       & sf EER (\%) & rf EER (\%)\\
    \midrule
    Trajectory image (single branch)    & 15.68 & 2.01\\
    GASF-only (single branch)           & 9.04$^{}$   & 2.37\\
    GADF-only (single branch)           & 8.42$^{}$   & 1.98\\
    Dual, concat only                   & 6.62$^{**}$       & 1.21$^{**}$\\
    \textbf{Dual, cross-attn (ours)}    & \textbf{6.01}$^{*}$ & \textbf{1.02}$^{*}$\\
    \bottomrule
  \end{tabular}
\end{table}

Three findings emerge from Table~\ref{tab:ablation_branch}.

\textbf{GADF is more discriminative than GASF alone} (8.42\% vs.\ 9.04\%). The anti-symmetric difference field $G^D[i,j] = \sin(\phi_i - \phi_j)$ encodes whether a kinematic value is rising or falling between two time steps; GASF's symmetric cosine sum instead encodes joint co-occurrence and cannot distinguish directional change. For a skilled forger who reproduces the gross speed profile but cannot replicate the precise acceleration/deceleration sequence, GADF's directional encoding is more discriminative.

\textbf{The trajectory image is the weakest single-branch configuration} (15.68\%), substantially worse than either GAF variant. Although the pen-path raster preserves 2D spatial shape, it discards all temporal structure — velocity, pressure dynamics, and directional transitions are invisible to the backbone. This confirms that temporal encoding, not spatial shape alone, is what drives discriminability against skilled forgeries.

\textbf{Dual-branch cross-attention outperforms all single-representation models} (6.01\%). Concatenation already recovers much of the benefit of combining both views (6.62\%), but bidirectional cross-attention adds a further 0.61\% by allowing each branch to attend to tokens in the other branch where the complementary representation is most discriminative, rather than fusing blindly at the embedding level.

\subsection{Embedding Space Analysis}
\label{sec:embedding}
To understand per-dataset performance variation, Table~\ref{tab:embedding} reports the mean cosine similarity between genuine--genuine pairs ($\mu_g$) and genuine--forged pairs ($\mu_f$) per sub-dataset, together with the margin $\Delta = \mu_g - \mu_f$. A larger margin indicates wider genuine/forgery separation in the learned metric space and directly predicts lower EER.
\begin{table}[!ht]
  \centering
  \caption{Embedding space analysis per sub-dataset. $\mu_g$\,=\,mean genuine--genuine cosine similarity; $\mu_f$\,=\,mean genuine--forged cosine similarity; $\Delta = \mu_g - \mu_f$.}
  \label{tab:embedding}
  \small
  \setlength{\tabcolsep}{5pt}
  \begin{tabular}{@{}lccc@{}}
    \toprule
    Dataset        & $\mu_g$ & $\mu_f$ & $\Delta$\\
    \midrule
    BiosecurID     & 0.860 & 0.261 & \textbf{0.599}\\
    MCYT           & 0.858 & 0.377 & 0.481\\
    BiosecureDS2   & 0.798 & 0.332 & 0.466\\
    eBioSignDS2    & 0.849 & 0.434 & 0.415\\
    eBioSignDS1    & 0.823 & 0.457 & 0.366\\
    \midrule
    DeepSignDB     & 0.836 & 0.351 & 0.485\\
    \bottomrule
  \end{tabular}
\end{table}
The margin $\Delta$ correlates directly with EER across all sub-datasets: BiosecurID ($\Delta{=}0.599$) achieves the lowest skilled EER (2.97\%), while eBioSignDS1 ($\Delta{=}0.366$) yields the highest (11.55\%). The reduced margin on eBioSignDS1 is driven by elevated forged similarity $\mu_f{=}0.457$: signatures collected across five working conditions introduce high intra-writer variability, causing genuine embeddings to spread and forged embeddings to fall closer to the genuine cluster. BiosecurID's low $\mu_f{=}0.261$ reflects clean pressure recordings that produce tight, well-separated genuine clusters, consistent with its strong per-dataset EER.

\subsection{Loss Components}

Table~\ref{tab:ablation_loss} ablates the training objective. A notable result is that adding forgery injection \emph{without} uniformity regularisation actually hurts performance (13.85\% vs.\ 11.74\% for triplet alone). Skilled forgeries share spatial trajectory structure with genuine samples and push the model toward hard decision boundaries, but without a spreading force on the hypersphere this causes embedding collapse: all genuine samples cluster in a small region, cosine scores become insensitive to writer identity, and the forgery triplets overfit to a degenerate solution. The uniformity regulariser~\cite{wang2020understanding} alone already improves over the baseline (10.37\%), and only when all three components are combined does the full benefit emerge (6.01\%): uniformity keeps the embedding space spread while forgery injection sharpens the genuine/forged boundary.

\begin{table}[!ht]
  \centering
  \caption{Loss component ablation, DeepSignDB ($R_\text{enroll}{=}4$, sf EER). Best in \textbf{bold}.}
  \label{tab:ablation_loss}
  \small
  \setlength{\tabcolsep}{4pt}
  \begin{tabular}{@{}ccc c@{}}
    \toprule
    Triplet & +Forgeries & +Uniformity & sf EER (\%)\\
    \midrule
    \checkmark &            &             & 11.74\\
    \checkmark & \checkmark &             & 13.85\\
    \checkmark &            & \checkmark  & 10.37\\
    \checkmark & \checkmark & \checkmark  & \textbf{6.01}\\
    \bottomrule
  \end{tabular}
\end{table}

The uniformity regulariser~\cite{wang2020understanding} prevents the common failure mode of embedding collapse, where all genuine samples cluster in a small region of the hypersphere and cosine scores become insensitive to writer identity. Without it, semi-hard triplet loss with skilled forgeries can overfit to the hardest pairs and produce a degenerate solution where random forgery EER improves but skilled forgery EER remains poor.

\subsection{Temporal Resolution}
\label{sec:abl_M}

The resample length $M$ controls the temporal resolution of the GAF images, which are computed at size $\frac{M}{2}{\times}\frac{M}{2}$ (Section~\ref{sec:gaf}). Table~\ref{tab:ablation_M} reports DeepSignDB ($R_\text{enroll}{=}4$) skilled EER as $M$ varies. The jump from $M{=}128$ to $M{=}256$ (10.55\% $\to$ 6.81\%) is the dominant transition: at $M{=}256$ the ConvNeXt-Tiny backbone produces a $4{\times}4$ spatial grid of 16 tokens, which appears sufficient for the self-attention head to aggregate discriminative temporal structure. At $M{=}64$ (a single token per branch), self-attention is vacuous and EER degrades to 10.72\%. All experiments use $M{=}512$.

\begin{table}[!ht]
  \centering
  \caption{Temporal resolution ablation, DeepSignDB ($R_\text{enroll}{=}4$, sf EER).}
  \label{tab:ablation_M}
  \small
  \setlength{\tabcolsep}{5pt}
  \begin{tabular}{@{}cccc@{}}
    \toprule
    $M$ & Image size & Tokens & sf EER (\%)\\
    \midrule
    64  & $32{\times}32$   & 1  & 10.72\\
    128 & $64{\times}64$   & 4  & 10.55\\
    256 & $128{\times}128$ & 16 & 6.81\\
    512 & $256{\times}256$ & 64 & \textbf{6.01}\\
    \bottomrule
  \end{tabular}
\end{table}

\subsection{Kinematic Channel Selection}
\label{sec:abl_channels}

Table~\ref{tab:ablation_channels} ablates the three kinematic channels independently and in pairs on the dual-branch model.

Speed $v$ is the most discriminative single channel (8.21\%), followed by pressure derivative $\dot{p}$ (9.77\%) and direction angle $\theta$ (10.37\%). Pairs consistently outperform singletons: $v{+}\dot{p}$ (7.43\%) is the strongest pair, closely approaching the full model, while $\dot{p}{+}\theta$ (8.04\%) and $v{+}\theta$ (8.63\%) also improve over their respective singletons. The full three-channel model (6.01\%) outperforms all pairs, confirming that speed, pressure derivative, and direction angle each carry complementary kinematic information.

\subsection{GAF Encoding Variant}
\label{sec:abl_gaf}

We compare our asymmetric GAF construction against a symmetric baseline, where the full time series is downsampled to ${M}$ points before computing the standard Gramian matrix, producing identical upper- and lower-triangle structure. Our asymmetric construction instead uses the first and second halves of the series as the two axes, preserving causal temporal ordering in the off-diagonal entries. On DeepSignDB ($R_\text{enroll}{=}4$) skilled EER, the symmetric variant yields 7.16\% versus 6.01\% for our asymmetric encoding, confirming that preserving temporal directionality provides a discriminative signal that is lost when the Gramian is forced to be symmetric.

\begin{table}[t]
  \centering
  \caption{Kinematic channel ablation, DeepSignDB ($R_\text{enroll}{=}4$, sf EER, dual-branch). Best in \textbf{bold}.}
  \label{tab:ablation_channels}
  \small
  \setlength{\tabcolsep}{5pt}
  \begin{tabular}{@{}lc@{}}
    \toprule
    Channels & sf EER (\%)\\
    \midrule
    Speed $v$ only                              & 8.21\\
    Pressure deriv.\ $\dot{p}$ only            & 9.77\\
    Direction $\theta$ only                     & 10.37\\
    \midrule
    $v + \dot{p}$ (no $\theta$)                & 7.43\\
    $\dot{p} + \theta$ (no $v$)                & 8.04\\
    $v + \theta$ (no $\dot{p}$)                & 8.63\\
    \midrule
    \textbf{All three} ($v + \dot{p} + \theta$) & \textbf{6.01}\\
    \bottomrule
  \end{tabular}
\end{table}

\subsection{Computational Complexity}
\label{sec:complexity}

GAF encoding adds a one-time $O(M^2)$ preprocessing step per signature, the same asymptotic cost as Transformer self-attention on a length-$M$ sequence. The resulting $(M/2){\times}(M/2)$ image is then processed by two ConvNeXt branches in $O(M^2)$ via fully parallelisable 2D convolutions, versus $O(M \cdot H^2)$ sequential steps for RNNs and $O(M \cdot C \cdot K \cdot L)$ for TCN/1D-CNN. The key distinction is representational: GAF encodes all pairwise temporal correlations \emph{explicitly} as pixel values, making global structure directly accessible to every convolutional layer, rather than requiring the network to re-discover it through depth as sequence models must.

\section{Conclusion and Future work}
\label{sec:conclusion}
We presented GAFSV-Net, a dual-branch writer-independent OSV framework encoding kinematic channels as asymmetric GASF/GADF images, processed by ConvNeXt-Tiny backbones linked via bidirectional cross-attention. GAFSV-Net outperforms DTW on BiosecurID and all sequence-based baselines under identical objectives, confirming that 2D temporal encoding captures structure 1D models miss and establishing GAF as a viable paradigm for OSV. This opens avenues for future work: applying saliency maps to GAF images to expose which temporal intervals and kinematic co-occurrences drive decisions, improving forensic interpretability; genuine-only training via self-supervised or one-class objectives to remove forgery dependence; and extending GAF to multiple temporal scales to capture fine local and coarse global dynamics.

{\small
\bibliographystyle{ieee}
\bibliography{egbib}
}

\end{document}